\documentclass[conference]{IEEEtran}
\usepackage{times}

\usepackage{amsmath}
\usepackage{amsfonts}
\usepackage{amssymb}
\usepackage[numbers]{natbib}
\usepackage{multicol}
\usepackage{graphicx}
\let\labelindent\relax
\usepackage{enumitem}
\usepackage{xcolor}
\usepackage{algorithm}
\usepackage{algpseudocode}

\newcommand{\rightcomment}[1]{\hfill\textcolor{gray}{\% #1}}

\usepackage[hypcap]{caption}
\usepackage{hyperref}
\usepackage{cleveref}
\hypersetup{
    colorlinks=true,
    linkcolor=blue,
    filecolor=magenta,
    urlcolor=cyan,
    pdftitle={Overleaf Example},
    pdfpagemode=FullScreen,
    }

\makeatletter
\def\blfootnote{\gdef\@thefnmark{}\@footnotetext}
\makeatother
\begin{document}
\title{Efficient Reinforcement Learning for Autonomous Driving with Parameterized Skills and Priors}

\author{\vspace{0.5em}Letian Wang$^1$, Jie Liu$^2$, Hao Shao$^3$, Wenshuo Wang$^4$, Ruobing Chen$^3$, Yu Liu$^{2,3,\dag}$, Steven L. Waslander$^1$\\
$^1$University of Toronto, $^2$Shanghai Artificial Intelligence Laboratory, $^3$SenseTime Research, $^4$McGill University}



%

\maketitle

\IEEEpeerreviewmaketitle

\begin{abstract}
When autonomous vehicles are deployed on public roads, they will encounter countless and diverse driving situations. Many manually designed driving policies are difficult to scale to the real world. Fortunately, reinforcement learning has shown great success in many tasks by automatic trial and error. However, when it comes to autonomous driving in interactive dense traffic, RL agents either fail to learn reasonable performance or necessitate a large amount of data. Our insight is that when humans learn to drive, they will 1) make decisions over the high-level skill space instead of the low-level control space and 2) leverage expert prior knowledge rather than learning from scratch. Inspired by this, we propose ASAP-RL, an efficient reinforcement learning algorithm for autonomous driving that simultaneously leverages motion skills and expert priors. We first parameterized motion skills, which are diverse enough to cover various complex driving scenarios and situations. A skill parameter inverse recovery method is proposed to convert expert demonstrations from control space to skill space. A simple but effective double initialization technique is proposed to leverage expert priors while bypassing the issue of expert suboptimality and early performance degradation. We validate our proposed method on interactive dense-traffic driving tasks given simple and sparse rewards. Experimental results show that our method can lead to higher learning efficiency and better driving performance relative to previous methods that exploit skills and priors differently. \href{https://github.com/Letian-Wang/asaprl}{Code} is open-sourced to facilitate further research.

\blfootnote{\dag  Corresponding author. liuyuisanai@gmail.com}
\blfootnote{First author contact: lt.wang@mail.utoronto.ca}

\end{abstract}
\section{Introduction}
\label{sec:intro}
Autonomous vehicles (AVs) on public roads will interact with other agents in various driving scenarios and situations characterized by traffic densities, road geometries, and traffic rules~\cite{wang2022social}. Many existing decision-making frameworks are based on elaborate hand-designed rules and decision hierarchies~\cite{ding2021epsilon,zhang2020efficient}. Nonetheless, the joint consideration of multiple vehicles scales exponentially in dense traffic and is usually computational resource-hungry. Further, it can be challenging to design rules manually to cover all safety-critical cases, leading to severe generalizability issues. Fortunately, reinforcement learning (RL) requires little human labor in identifying policies across multiple different tasks by automatically interacting with the environment ~\cite{silver2016mastering,vinyals2019grandmaster,liu2022motor}. However, when it comes to an interactive multi-vehicle setting with a continuous action space, the learning efficiency of RL algorithms remains notoriously low because RL agents either fail to learn reasonable performance or necessitate a large amount of data and resources to make significant progress.

\begin{figure}[!t]
    \begin{center}
    \includegraphics[width=0.48\textwidth]{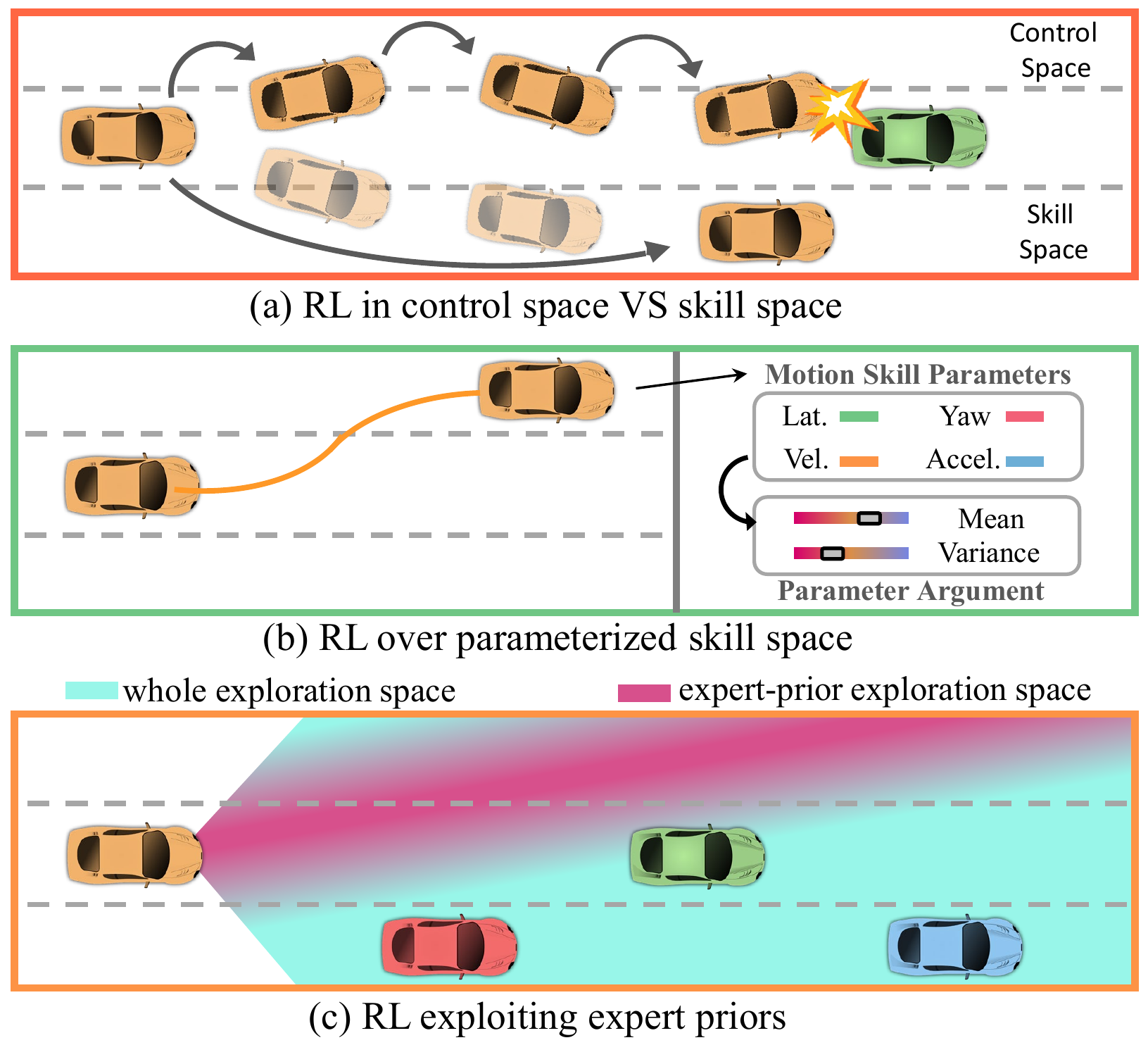}
    \end{center}
    \caption{(a) RL-based AVs learning over the control space will exhibit inconsistent action sequences. In comparison, RL over the skill space can generate a sequence of consistent low-level actions with more informative exploration and accelerated reward signaling. (b) The parameterized motion skill provides a key interface for RL agents to explore and learn. (c) The expert demonstration can provide prior knowledge of which regions of the action space are more promising in getting rewards than others, which can accelerate learning.}
    \label{fig: motivation}
    \vspace{-1em}
\end{figure}

%

One great insight to improve learning efficiency is that, there can be different choices of the action space for the RL agent, and a proper choice or design of the action space can significantly simplify learning~\cite{martin2019variable}. Most existing RL methods directly learn over vehicles' control space, such as the steering and the pedal commands of one time step ~\cite{sharifzadeh2016learning,chen2021interpretable,peng2021learning}. However, a sequence of single-step control signals with inconsistent exploration rarely achieves typical driving maneuvers and lacks informative reward signals for the agent to learn.
For example, as shown in the upper part of Fig.~\ref{fig: motivation}(a), the orange vehicle might move erratically when exploring the control space, thereby failing to achieve driving maneuvers such as overtaking a leading car. Such agents with frequent failures rarely receive informative reward signals for the agent to improve. In contrast, behavioral science has revealed that human behavior is naturally temporally hierarchical ~\cite{jing2004construction,botvinick2009hierarchically,BerliacHierachialRL2019}: humans learn and make decisions over abstracted high-level options, which we call motion skills, while the low-level control commands are not an action space to learn but simply a sequence of muscle responses/executions to achieve high-level motion skills. As shown in the lower part of Fig.~\ref{fig: motivation}(a), a sequence of consistent low-level control commands is generated correspondingly after the driver decides to perform the motion skill of overtaking a lead vehicle. Such temporally-extended motion skills enable efficient learning through structured exploration and improved reward signaling.

However, proper design and learning of motion skills are non-trivial. There are two main ways to define and learn the motion skills in autonomous driving and robotics: (1) Manually designing delicate task-specific or object-centric motion skills ~\cite{wang2022transferable,deo2018would,dalal2021accelerating,wang2021learning,kober2009learning,peters2008reinforcement,wang2021hierarchical}, such as merging into a target lane before a car. 
However, such task-specific or object-centric motion skills are usually too complex to design manually for autonomous driving in multi-agent dense-traffic settings, nor can such delicately-designed motion skills cover the wide variety of driving and interaction situations that arise. 
(2) Extracting or learning motion skills from offline motion datasets, such as clustering or segmenting motion trajectories ~\cite{konidaris2012robot,rao2021learning,pertsch2020keyframing}, or distilling offline motions into low-dimension latent space ~\cite{pertsch2020accelerating,merel2018neural,merel2020catch}. However, motion datasets are usually unbalanced in distribution and lack diversity, making it hard to learn all the needed skills. Inspired by motion planning in autonomous driving, we propose to exploit motion skills in the pure ego vehicle motion view, which is diverse and thus generalizable to complex driving tasks. Fig.~\ref{fig: motivation}(b) depicts that with such naturally-parameterized motion skills, AVs can directly learn over the parameters of motion skills with little design effort.



In addition to using motion skills, the other widely-recognized attempt to improve learning efficiency is leveraging prior knowledge from expert demonstrations. The critical insight is that the value of the whole exploration space is not uniform. Some regions of the action space are more promising in getting rewards than others. As shown in Fig.~\ref{fig: motivation}(c), it would be more rewarding for the vehicle to run in the left-front direction with sparse traffic than in the right-front direction with dense traffic. Many works integrate such expert prior knowledge by utilizing expert demonstrations to initialize the RL agent  ~\cite{Liu2022ImprovedDR,hester2018deep,liang2018cirl,nair2020awac,rajeswaran2017learning} and/or training an expert policy based on expert demonstrations to guide reinforcement learning as a regularizer or reward term~\cite{pertsch2020accelerating,rengarajan2022reinforcement,liu2022motor}. 
However, simultaneously leveraging the expert's prior knowledge and the parameterized motion skills is non-trivial, since most expert demonstrations only include control information but miss skills or rewards annotation. To this end, we propose an inverse optimization method to recover the corresponding motion skill parameters given expert demonstrations using sequential quadratic programming (SQP)~\cite{kraft1988software}. Thus, we can convert the expert demonstration from control space to skill space, 
allowing us to take advantage of both skills and priors simultaneously. 

While actor pertaining ~\cite{Liu2022ImprovedDR,hester2018deep,liang2018cirl,nair2020awac,rajeswaran2017learning} and expert regularization~\cite{pertsch2020accelerating,rengarajan2022reinforcement,liu2022motor} have been widely used in previous methods to leverage expert priors, such methods could either suffer from performance drop at early training iterations due to actor/critic mismatch or performance suppression due to expert suboptimality. To this end, we propose a simple but effective double initialization method. We first pretrain the actor with the expert demonstration in skill space, then pretrain the critic by rolling out the pretrained actor to collect expert demonstrations with reward and skill information. We will show how this simple but effective double initialization method can bypass the issue of suboptimality and initial performance drop.


In summary, the contributions of the proposed ASAP-RL (RL with p\textbf{A}rameterized \textbf{S}kills \textbf{a}nd \textbf{P}riors) are threefold:
\begin{itemize}
    \item Propose an RL method to learn over the parameter of motion skills for more informative exploration and improved reward signaling. Such skills are defined in the ego vehicle motion view, which are diverse and thus generalizable to different complex driving tasks. 
    \item Propose an inverse skill parameter recovery method to convert expert demonstration from control space to skill space, and a simple but effective double initialization method to better leverage expert prior without issues of performance drop or suppression. Thus we can take advantage of both skills and priors simultaneously.
    \item Validate our method for autonomous driving tasks in three challenging dense-traffic scenarios and demonstrate our method outperforms previous methods that consider skills and priors differently.
\end{itemize}




\section{Related Works}
\label{sec:related}

\begin{figure*}[!t]
    \begin{center}
    \includegraphics[width=0.9\textwidth]{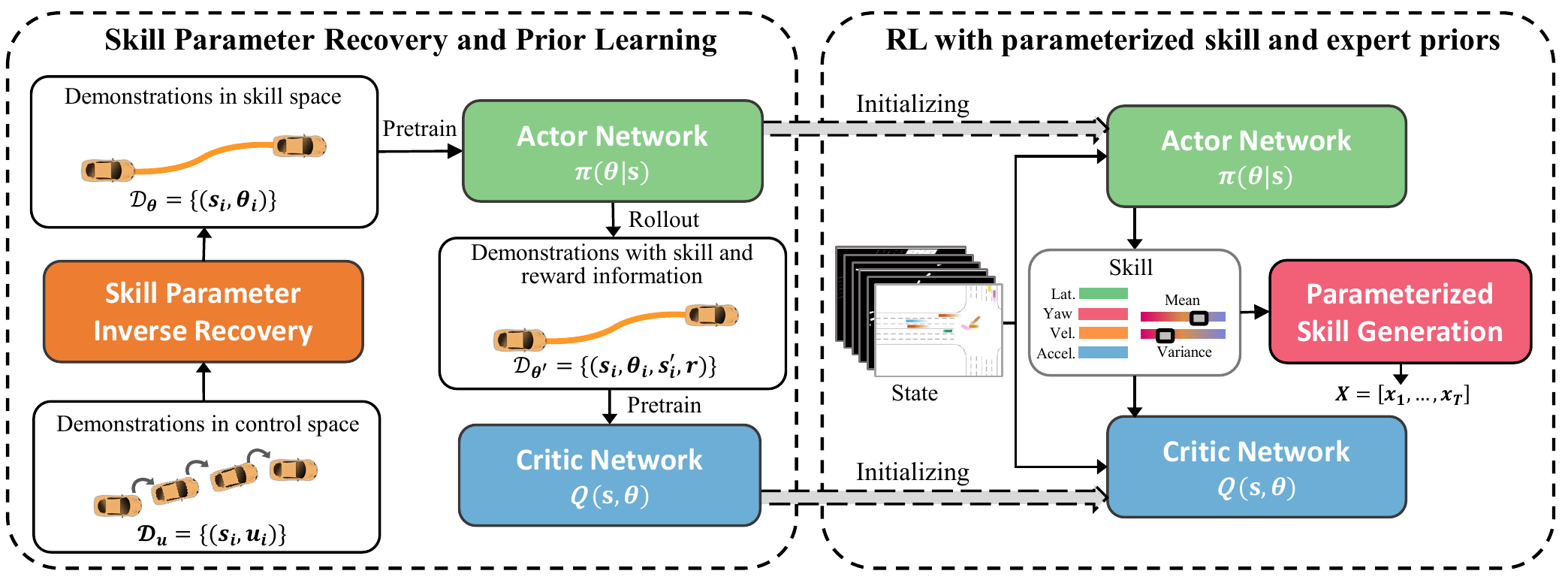}
    \end{center}
    \caption{The pipeline of the proposed ASAP-RL method. An inverse skill parameter recovery method is proposed to convert expert demonstration from control space to skill space. A double initialization method is introduced to initialize both actor and critic to inject the expert's prior knowledge into RL. The RL agent can learn and explore in the skill space instead of the control space while leveraging the expert priors, which leads to high learning efficiency and improved final performance.}
    \label{fig: main_graph}
    \vspace{-1em}
\end{figure*}
In this section, we will discuss related works on reinforcement learning with skills and priors, with a focus on autonomous driving and robotics. To the best of our knowledge, we are the very first to simultaneously leverage skills and expert priors to improve reinforcement learning for autonomous driving. Moreover, our proposed skill representation is motion-centric, not task-centric, implying that it can be flexibly extended to navigation for other moving robots, such as for mobile robots, UAVs, and the end-effector of manipulation robots, by accordingly modifying the dynamic constraint of the skill. Our proposed skill recovery and double-initialization method is flexible to leverage prior knowledge for general robot embodiment. We believe the proposed method could help to close the gap between skill and priors, and spark further research in this direction.

\subsection{Reinforcement Learning with Skills}
Toward defining or getting the skill that can be exploited in reinforcement learning for autonomous driving and robotics, two main approaches exist: 
(1) manually designing task-specific or object-centric motion skills~\cite{wang2022transferable,deo2018would,dalal2021accelerating,wang2021learning,kober2009learning,peters2008reinforcement}, such as cutting in a target lane before a specific car, or grasping a specific object. However, when AVs run on actual roads, a dense-traffic setting, AVs' motion usually should be synthesized considering relationships between multiple surrounding vehicles, which is usually too complicated to design manually. With limited flexibility and expressiveness, such manually-designed skills can hardly cover diverse driving and interaction situations. 
(2) another approach is to extract or learn skills from offline motion datasets, such as clustering or segmenting motion skills~\cite{konidaris2012robot,rao2021learning,pertsch2020keyframing}, or distill offline motions into low-dimension latent space~\cite{pertsch2020accelerating,merel2018neural,merel2020catch}, so that the issues and labor of manual design can be bypassed. However, such datasets can be expensive and labor-intensive to collect if they are not already available. It would also be difficult for such learned skills to transfer to new tasks or environments as they are usually task-specific and conditioned on the environment. Moreover, it is also not guaranteed that all necessary skills are covered in the datasets as they are usually unbalanced in distribution and sub-optimal. 
To tackle the limitations above, one recent work~\cite{zhou2022accelerating} combines the two approaches by first building a task-agnostic and ego-centric motion skill library in a pure ego vehicle motion perspective and then encoding the motion skills into a low-dimension latent skill space. However, the skill library construction still requires considerable effort, and the latent space encoding further makes the decision-making less interpretable. Thus, in this work, we propose to directly learn over the skill parameter space instead of the latent space, to spare the efforts in skill library construction, enable more interpretability in the decision-making process, and also provide an parameterized interface to further leverage priors.

\subsection{Reinforcement Learning with Expert Priors}
\label{related work: prior}
Although RL has been shown to be effective in several problems, learning efficiency issues~\cite{jin2021bellman} limit its applications. Inspired by human decision-making processes where prior knowledge is quite helpful when we learn new tasks, many works use expert prior knowledge to avoid learning from scratch with exhaustive interactions. There exist three approaches to leveraging prior knowledge:
(1) utilize expert demonstration for a warm-up pretraining or policy (actor) initialization before RL~\cite{Liu2022ImprovedDR,hester2018deep,liang2018cirl,nair2020awac,rajeswaran2017learning}; (2) train an expert policy based on expert demonstrations, which is then used to guide RL process~\cite{pertsch2020accelerating,rengarajan2022reinforcement,liu2022motor}; (3) maintain an expert data buffer, which is mixed with interaction buffer during RL for more fruitful experiences~\cite{Liu2022ImprovedDR,hester2018deep,nair2020awac,rajeswaran2017learning,vecerik2017leveraging,nair2018overcoming,lu2022imitation}. The first method is effective in policy-based RL methods but only initializing the actor was reported to be not helpful in the actor-critic framework~\cite{pertsch2020accelerating}. 
This is because the actor and critic are interacting with each other during the learning process, and the actor's objective is to maximize the Q-value output by the critic without pretrianing. Thus, the actor could quickly lose the prior knowledge learned from the expert after several updates. Empirically, we found performance drops can happen at early training iterations when adopting the first method (see Section~\ref{abl: prior}). The second and third methods use expert priors as guidance during RL training but might suppress the performance when the expert performance is suboptimal~\cite{rengarajan2022reinforcement}. To overcome the limitations of the three methods, we propose a `double initialization' technique with both the actor and critic initialized simultaneously and demonstrate this approach can achieve strong performance even when a suboptimal expert is employed.

\section{Approach}
Let the Markov Decision Process be defined as $\{\mathcal{S}, \mathcal{A}, \mathcal{T}, \mathcal{R}, \gamma\}$, a tuple of states, actions, transition probabilities, rewards, and discount factor. 
Our goal is to leverage parameterized motion skills and priors to accelerate reinforcement learning in continuous space for autonomous driving in dense traffic settings.
Fig.~\ref{fig: main_graph} illustrates the architecture of our method, and Algorithm~\ref{alg:asaprl} outlines the pipeline. In this section, we first define the parameterized motion skill, which is defined in pure ego vehicle motion view and thus generalizable to different driving scenarios.
Then the RL agent only needs to focus on the high-level decision-making task by learning over the skill parameter space, and the low-level motion control task is handled by generating the motion trajectory from the skill parameters. To further utilize priors in expert demonstrations $\mathcal{D}_u=\{(s_i, u_i)\}$, which only include state $s_i$ and control action $u_i$ but lack skill or reward information, we proposed an inverse optimization process to recover skill parameters $\boldsymbol{\theta}_i$, constructing the expert demonstrations in skill space $\mathcal{D}_{\boldsymbol{\theta}}=\{(s_i, \boldsymbol{\theta}_i)\}$. In a maximum entropy actor-critic framework, we propose a simple but effective `double initialization' method to pretrain both the actor and critic, which can bypass the issue of suboptimality and early performance drop when leveraging experts prior. With these methods, we can leverage the parameterized skill and prior simultaneously. 

\subsection{Motion Skill Generation}
\label{sec:skill generation}
\begin{figure}[!t]
    \begin{center}
    \includegraphics[width=0.45\textwidth]{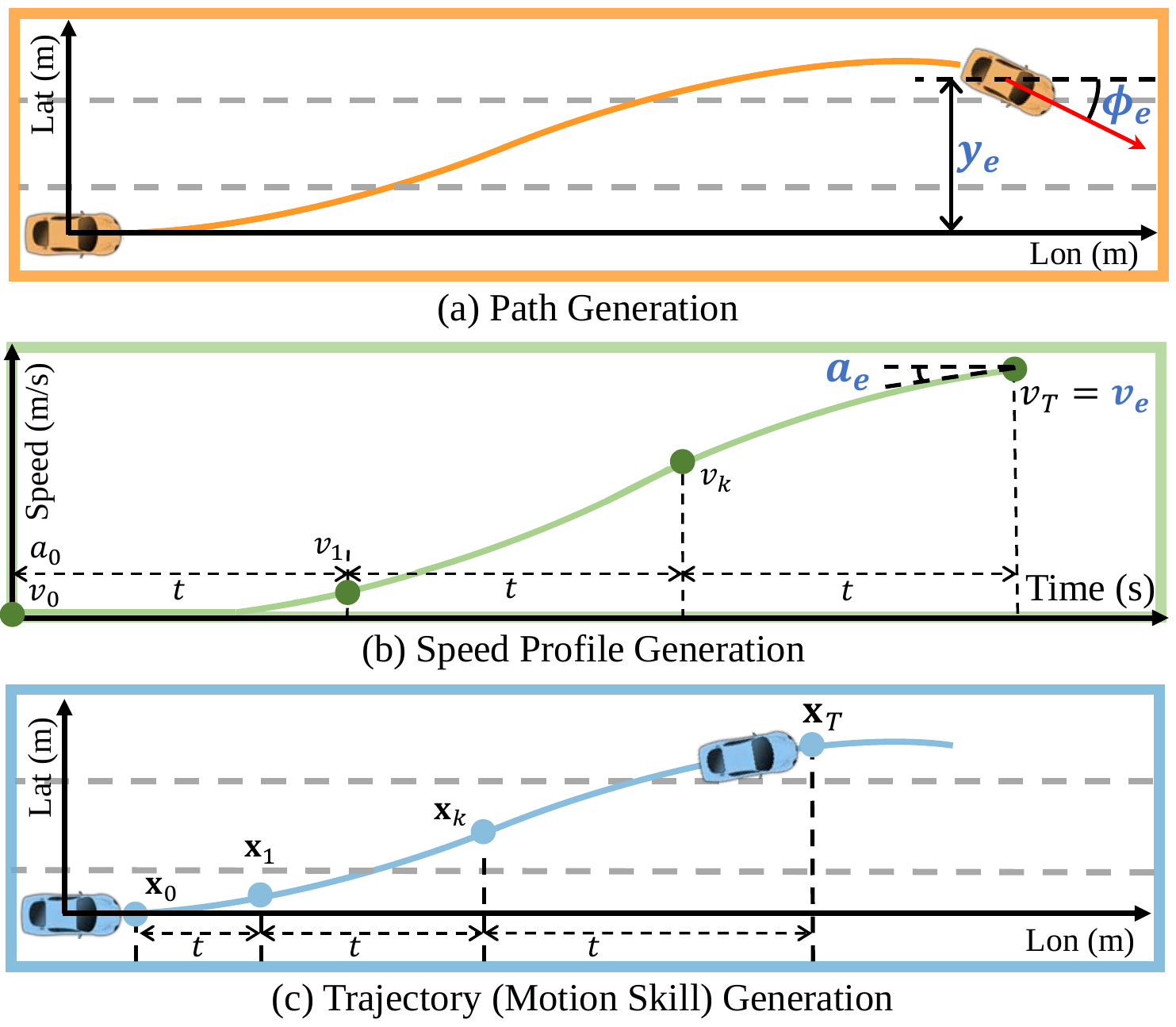}
    \end{center}
    \caption{An illustration of parameterized motion skill generation process. One motion skill is determined by four skill parameters (shown in blue color) that RL agents directly learn and explore. (a) The path is generated by connecting a start point and an endpoint (parameterized by the lateral position \textcolor{blue}{$y_e$} and heading angle \textcolor{blue}{$\phi_e$} of the endpoint) by the cubic polynomial. (b) The speed profile is represented by a cubic polynomial within the time window $T$, which is parameterized by the speed $v_0$ and acceleration $a_0$ at the beginning time and \textcolor{blue}{$v_e$} and \textcolor{blue}{$a_e$} at the end time. (c) Motion skill generated by projecting the integral of the speed profile onto the path.}
    \label{fig: motion}
    \vspace{-1em}
\end{figure}

Inspired by the sampling-based motion planning~\cite{li2020IFAC,gu2017improved,wang2021socially}, we first introduce the parameterized motion skill. Defined in a pure ego-motion perspective, such motion skills are task-agnostic and ego-centric and can cover diverse motions needed in dense-traffic settings, which can be generalizable to different scenarios. Technically, the beginning boundary of the motion skill is determined by the vehicle’s current state, and one motion skill further necessitates four parameters of the end boundary i.e., the lateral position $y_e$, heading angle $\phi_e$, speed $v_e$, and acceleration $a_e$ at the end of the motion skill. RL agents can learn and explore these parameters to generate diverse motion skills.
Given the skill parameters, the generation of one motion skill consists of three steps, as shown in Fig.~\ref{fig: motion}: (a) path generation, generating future path on the road, (b) speed profile generation, specifying the variation of speed in the skill time window, and (c) trajectory (motion skill) generation, projecting the integral of the speed profile onto the path to generate motion trajectory. All details are introduced below.

\textbf{Path Generation.} The path is generated by connecting the origin and an endpoint on the road by cubic splines, in the ego vehicle's coordinate system. The endpoint is characterized by three parameters: longitudinal position $x_e$, lateral position $y_e$, and heading angle $\phi_e$. To ensure feasible speed-to-path projection, the path length should be longer than the integral of the speed profile. We thus fix the longitudinal position of the endpoint as the longest distance the ego vehicle can reach within the skill time window $T$.
Therefore, the path generation consists of two free parameters the RL agent needs to learn: the lateral position $y_e$ and heading angle $\phi_e$ of the endpoint, which can cover diverse lateral driving intentions and maneuvers, such as lane keeping, overtaking, and cutting-in. 

\textbf{Speed Profile Generation.}
The speed profile is represented by a cubic polynomial in the time horizon $[0, T]$, parameterized by the speed $v$ and acceleration $a$ at the beginning and end of the time horizon. When generating the speed profile, we specify the speed and acceleration at the beginning time by the vehicle's current state. Thus, the speed profile generation phase provides two free parameters that the RL agent needs to learn: the speed $v_e$ and acceleration $a_e$ at the end time, which can cover diverse temporal intentions, such as accelerating, decelerating, and emergent stop.

\textbf{Parameterized Motion Skill Generation.}
Given the path and speed profile, the motion skill is generated by projecting the integral of the speed profile onto the path. Each motion skill is a sequence of vehicle states $\mathbf{X} = [\mathbf{x_1}, \mathbf{x_2}, ..., \mathbf{x_{T}}]$, with each state as a tuple $\mathbf{x_t} = \{x_t, y_t, \phi_t, v_t, a_t\}$. 

Note that each skill generation is conditioned on the vehicle’s current state, which is the final state of the last executed skill, which ensures smoothness between skill segments. The dynamic constraint (acceleration, curvature) is also enforced by restricting the planning parameters in reasonable ranges. 


\subsection{Skill Parameter Recovery}
\label{sec: prior learning}
Another way to accelerate RL is by leveraging prior knowledge from expert demonstrations. 
However, most expert demonstrations $\mathcal{D}_u=\{(s_i, u_i)\}$ are in control space and do not contain the skill and reward information, making it not readily usable. We propose an inverse parameter recovery procedure to annotate expert data with skill parameters and construct the expert demonstration in skill space $\mathcal{D}_{\boldsymbol{\theta}}=\{(s_i, \boldsymbol{\theta}_i)\}$.

Consider the motion skill generation (see Section~\ref{sec:skill generation}) as a forward process, where one motion skill $\mathbf{X}$ is generated given skill parameters $\boldsymbol{\theta}$. Then the skill parameter recovery can be regarded as an inverse procedure, where $\boldsymbol{\theta}$ are determined given one demonstrated motion skill  $\mathbf{X}_d$. Practically, the motion skills are retrieved by sequentially splitting the demonstration trajectory into segments, each with a length of $T$. Formally, this recovery process can be formulated as
\begin{align}
\begin{aligned}
    \boldsymbol{\theta} = & \arg\min_{\boldsymbol{\theta}} ||\mathbf{X}_d - \mathbf{X}||_2\\ \quad
    \mbox{s.t.} & \quad \mathbf{X} = f_s(\boldsymbol{\theta}) \\
\end{aligned}
\label{eq:recovering}
\end{align}
where $f_s$ denotes the motion skill generation process in Section~\ref{sec:skill generation}. Specifically, we use the sequential quadratic programming (SQP) ~\cite{kraft1988software} as the optimization method. For each motion skill, we run the optimization process multiple times with different skill parameter initializations  to mitigate issues of local optima and achieve better recovery accuracy.

\subsection{Expert Prior Learning - Actor and Critic Pretraining}
With the expert demonstration in skill space $\mathcal{D}_{\boldsymbol{\theta}}=\{(s_i, \boldsymbol{\theta}_i)\}$, we can leverage the skill and expert prior simultaneously in RL. In this paper, we modify the maximum-entropy RL to do so. Specifically, we adopt the Soft Actor-Critic framework (SAC~\cite{haarnoja2018soft}), which consists of an actor $\pi(\boldsymbol{\theta}|s)$ for policy improvement and a critic $Q(s,\boldsymbol{\theta})$ for policy evaluation. This section will introduce how the two networks can be pretrained to capture expert priors.

\textbf{Actor Pretraining} To leverage the expert demonstration in skill space $\mathcal{D}_{\boldsymbol{\theta}}$ as priors, we first pretrain an actor $\pi(\boldsymbol{\theta}|s)$ to capture the skill priors conditional on the current state. The training of this model aims at maximizing the log probability of the recovered expert skill parameter in the distribution output by the actor:
\begin{equation}\label{eq:ELBO}
    \mathbb{E}_{(s,\boldsymbol{\theta}) \sim\mathcal{D}_{\boldsymbol{\theta}}} \bigg[\log \pi(\boldsymbol{\theta} \vert s) + \beta \mathcal{H}(\boldsymbol{\theta})\bigg]
\end{equation}
where for each data $(s, \boldsymbol{\theta})$ in the expert demonstration $\mathcal{D}_{\boldsymbol{\theta}}$, the pretrained actor $\pi(\boldsymbol{\theta}|s)$ takes the current state $s$ as inputs and outputs the Gaussian distribution of the skill parameters $\boldsymbol{\theta}$. $\mathcal{H}(\boldsymbol{\theta})$ denotes the entropy regularization term and $\beta$ denotes entropy weight. The pretrained actor can provide prior knowledge on which skills are more promising to explore conditioning in the current situation. 

\textbf{Critic Pretraining} Since the actor and critic interact with each other during RL training, only pretraining the actor does not fully leverage the prior knowledge. 
However, pretraining the critic $Q(s,\boldsymbol{\theta})$ is not always available since there is no reward information in the expert demonstration in either control space $\mathcal{D}_u$ or skill space $\mathcal{D}_{\boldsymbol{\theta}}$. Fortunately, we already have the pretrained actor who has learned expert priors. Thus we propose to roll out the pretrained actor in the environment to collect an expert demonstration with both skill and reward information $\mathcal{D}_{\boldsymbol{\theta}}^{'}=\{(s_i, \boldsymbol{\theta}_i, s_i', r)\}$, which is then used to pretrain the critic $Q(s,\boldsymbol{\theta})$. The pretaining of the critic follows typical policy evaluation in SAC, as in Line 38 of Algorithm~\ref{alg:asaprl}. As in Sec~\ref{abl: prior}, we will discuss in detail how the simple but effective double initialization method can outperform other methods to incorporate expert priors.

\begin{algorithm}[ht!]
\begin{algorithmic}[1]
\caption{ ASAP-RL} %
\label{alg:asaprl}
\State \textbf{Input:} Raw demonstrations $\mathcal{D}_u$, discount $\gamma$, target entropy $\delta$, learning rates $\lambda_{\pi}, \lambda_Q, \lambda_\alpha$, target update rate $m$, temperature hyperparameter $\alpha$, motion skill model $f_s$
\State \textbf{Require:} actor $\pi_\varphi(\boldsymbol{\theta}_t\vert s_t)$, critic $Q_\phi(s_t, \boldsymbol{\theta}_t)$, target network $Q_{\bar{\phi}}(s_t, \boldsymbol{\theta}_t)$, replay buffer $\mathcal{D}$, demonstration in skill space $\mathcal{D}_{\boldsymbol{\theta}}$, demonstration with skill and reward information $\mathcal{D}_{\boldsymbol{\theta}}^{'}$.
\State \textcolor{blue}{Skill Parameter Recovery}
\For{each trajectory in $\mathcal{D}_u$}
\For{every $\mathbf{X}_d$ split from the trajectory}
\State $\boldsymbol{\theta} = \arg\min ||\mathbf{X}_d -  f_s(\boldsymbol{\theta})||_p$ (Eq~\ref{eq:recovering})
\State $\mathcal{D}_{\boldsymbol{\theta}} \leftarrow \mathcal{D}_{\boldsymbol{\theta}} \cup \{(s, \boldsymbol{\theta})\}$ 
\EndFor
\EndFor
\State \textcolor{blue}{Prior Learning}
\For{each iteration}\rightcomment{Actor Pretraining}
\State Sample $(s, \boldsymbol{\theta})$ from $\mathcal{D}_{\boldsymbol{\theta}}$
\State Update $\pi_\varphi$ according to Eq~\ref{eq:ELBO}
\EndFor
\For{each iteration} \rightcomment{Roll-out $\pi_\varphi$ to Collect $\mathcal{D}_{\boldsymbol{\theta}}^{'}$}
\For{every $T$ environment step}
\State rollout pretrained actor to collect $\{s_t, \boldsymbol{\theta}_t, \tilde{r}, s_{t^\prime}\}$
\State $\mathcal{D}_{\boldsymbol{\theta}}^{'} \leftarrow \mathcal{D}_{\boldsymbol{\theta}}^{'} \cup \{s_t, \boldsymbol{\theta}_t, r, s_{t^\prime}\}$
\EndFor
\EndFor
\For{each iteration}\rightcomment{Critic Pretraining}
\State Sample $(s, \boldsymbol{\theta}, r, s')$ from $\mathcal{D}_{\boldsymbol{\theta}}^{'}$
\State Update $\pi_\varphi$ according to Line 38
\EndFor
\State \textcolor{blue}{RL with Parameterized Skills and Priors}
\State Initialize actor and critic with the pretrained weight
\For{each iteration}
\For{every $T$ environment step}
\State $\boldsymbol{\theta}_t \sim \pi_\varphi(\boldsymbol{\theta}_t \vert s_t)$ \rightcomment{sample skill parameter}
\State $\textbf{X}_t = \{\mathbf{x_i}\}^T_{i=1}\sim f_s(\textbf{X}_t \vert \boldsymbol{\theta}_t)$ \rightcomment{generate skill}
\State $s_{t^\prime} \sim p(s_{t+T},r_{t:t+T} \vert s_t, \textbf{X}_t)$ \rightcomment{execute skill}
\State $\tilde{r}_t(s_t, \boldsymbol{\theta}_t) = \sum_{i=0}^{T-1}r_{t+i}$ \rightcomment{reward calculation} 
\State $\mathcal{D} \leftarrow \mathcal{D} \cup \{s_t, \boldsymbol{\theta}_t, \tilde{r}, s_{t^\prime}\}$ \rightcomment{replay buffer}
\EndFor
\For{every gradient step} \rightcomment{typical SAC training}
\State \resizebox{.82\hsize}{!}{$\bar{Q} = \tilde{r}(s_t, \boldsymbol{\theta}_t) + \gamma \big[ Q_{\bar{\phi}}(s_{t^\prime}, \pi_\varphi(\boldsymbol{\theta}_{t^\prime} \vert s_{t^\prime})) - \alpha\mathcal{H}\big(\pi_\varphi(\boldsymbol{\theta}_{t^\prime} \vert s_{t^\prime})\big)\big]$} 
\State \resizebox{.82\hsize}{!}{$\varphi\leftarrow\varphi+\lambda_\pi\nabla_\varphi[Q_{{\varphi}}(s_{t}, \pi_\varphi(\boldsymbol{\theta}_{t}\vert s_{t}))+\alpha \mathcal{H}\big(\pi_\varphi(\boldsymbol{\theta}_{t}\vert s_{t})\big)]$} 
\State $\phi \leftarrow \phi - \lambda_Q \nabla_\phi \big[ \frac{1}{2}\big(Q_\phi(s_t, \boldsymbol{\theta}_t) - \bar{Q} \big)^2 \big]$ \label{alg:line:critic}
\State $\alpha \leftarrow \alpha - \lambda_\alpha \nabla_\alpha \big[ \alpha \cdot ((\mathcal{H}\big(p_{\boldsymbol{\theta}}(\boldsymbol{\theta}_{t} \vert s_{t}) - \delta) \big]$ 
\State $\bar{\phi} \leftarrow m \phi + (1 - m) \bar{\phi}$ 
\EndFor
\EndFor
\end{algorithmic}
\end{algorithm}

\subsection{RL over parameterized skill with priors}
To simultaneously leverage parameterized motion skill and expert priors for higher learning efficiency, we modify the maximum-entropy RL, whose objective function is to encourage reward maximization and exploration in skill space:
\begin{equation}
    J = \mathbb{E}_\pi \bigg[\sum_{t=1}^{T} \gamma^t r_t + \alpha\mathcal{H}\big(\pi(\boldsymbol{\theta}|s)\big)\bigg]
\end{equation}
where $\sum_{t=1}^{T} \gamma^t r_t$ denotes the accumulated discounted reward return from the environment after one motion skill of length $T$ is executed, $\mathcal{H}\big(\pi(\boldsymbol{\theta}|s)\big)$ denotes the entropy term, and $\alpha$ denotes the temperature parameter. Instead of learning a policy over raw control actions $\pi(u|s)$ at a single time step, we learn a policy that outputs skill parameters, $\pi(\boldsymbol{\theta}|s)$, which is then used to generate a motion skill by the procedure defined in Section~\ref{sec:skill generation}. Each motion skill is tracked for $T$ time steps before the next skill is generated, which follows a typical semi-MDP process~\cite{sutton1999between,bacon2017option} with temporal abstraction and accelerated reward signaling~\cite{merel2018neural,merel2020catch,liu2022motor}. The time horizon, $T$, is fixed and consistent with the motion skill length in Section~\ref{sec:skill generation} and Section~\ref{sec: prior learning}. While some works investigated variable-length policy conditional on the task and environment ~\cite{pertsch2020keyframing,kipf2018compositional,shankar2019discovering}, we empirically found a fixed-length policy can achieve strong performance and leave the extension to dynamic-length policies to future work.
On top of parameterized motion skills, we adopt the double initialization method to further leverage priors, where we initialize the actor and the critic with the pretrained weight in Section~\ref{sec: prior learning}.  

\section{Experiments}
In this section, we will investigate the following sub-problems to evaluate our proposed ASAP-RL method:
\begin{itemize}[leftmargin=*]
 \item \textbf{Performance}: Can our method learn driving strategies with higher learning efficiency and performance than other methods considering skills and priors differently? (Fig.~\ref{fig: main_result})
\end{itemize}
\begin{itemize}[leftmargin=*]
 \item
\textbf{Influence of the length of skill}:
The length of motion skill $T$ is an important hyper-parameter. How does it influence the performance of our ASAP-RL? (Fig.~\ref{fig: length})
\end{itemize}
\begin{itemize}[leftmargin=*]
 \item
\textbf{Influence of expert prior}: Does the expert prior knowledge accelerate RL? How does the proposed double initialization perform with respect to other methods to incorporate expert prior? (Fig.~\ref{fig: prior})
\end{itemize}

\subsection{Experiment Setup}
\subsubsection{Environment}
\begin{figure}[!t]
    \begin{center}
    \includegraphics[width=0.48\textwidth]{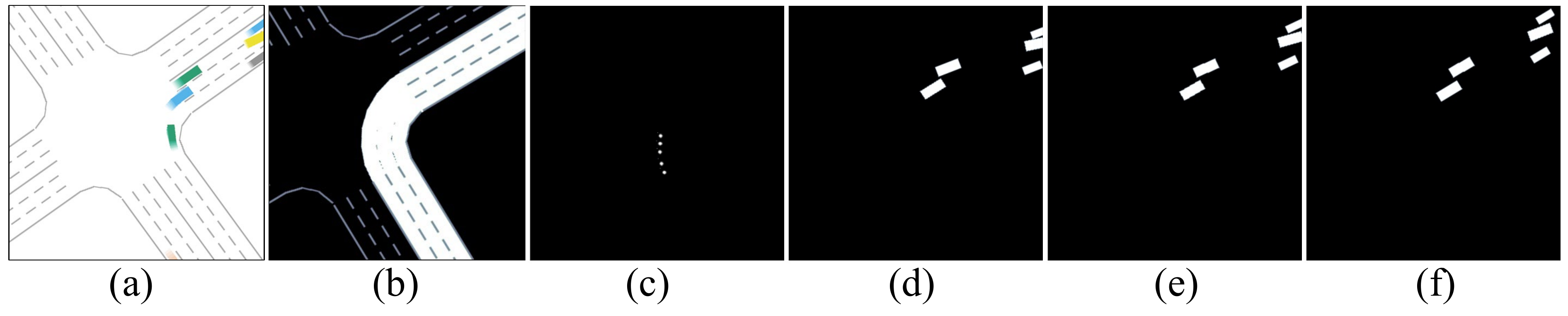}
    \end{center}
    \caption{The birds-eye view (BEV) images used as the observation and policy input. (a) the current scene; (b) road information (dashed line) and navigation lanes (in white color); (c) historical waypoints of the ego vehicle; (d-f) surrounding objects (white rectangles) at time $t$, $t-1$, and $t-2$.}
    \label{fig: obs}
    \vspace{-1em}
\end{figure}

We evaluate ASAP-RL on the MetaDrive simulator~\cite{li2022metadrive} under different dense-traffic scenarios (highway, roundabout, and intersection) to verify its performance for autonomous driving. At each run, the driving environment is generated with a random lane number and a random order of roadblocks. Traffic vehicles are spawned at a random location with a random target speed, and these vehicles are controlled by the default rule-based planner in MetaDrive. The ego vehicle needs to navigate in the traffic and arrive at the destination within the required time, without collisions or running off the roadway. As shown in Fig.~\ref{fig: obs}, we follow the default setting in MetaDrive to use a 5-channel birds-eye-view (BEV) image with a size of $200\times200\times5$ as the input for the RL agent, which includes spatiotemporal information of the ego agent and surrounding agents, and the information of the road geometry and navigation. In practice, such observations are usually available in modern autonomous driving perception systems\cite{shao2023safety}. The focus of this paper is how to make decisions efficiently and safely in diverse dense-traffic scenarios with access to these observations. All experiments are run with three different seeds.

\subsubsection{Reward Definition}s
We adopt the sparse reward setting to enable minimum effort in reward engineering:
\begin{equation}
    r_t = R_{progress} + R_{destination} + R_{crash} + R_{overtaking}.
\end{equation}
\begin{itemize}[leftmargin=*]
 \item  $R_{progress}$: The agent gets a sparse reward of 1 for every 10 m distance completed.
 \item  $R_{destination}$: The agent gets a reward of 1 if it reaches the destination.
 \item  $R_{crash}$: If the agent collides with other vehicles or the road curbs, it gets a negative reward of -5.
  \item  $R_{overtaking}$: If the agent passes one vehicle, it gets a reward of 0.1.
\end{itemize}

In the ASAP-RL pipeline, a motion skill step contains $T$ simulation steps, where each step lasts for 0.1 seconds. When the simulator performs one motion skill, it performs $T$ simulation steps and gets rewards $T$ times. Therefore, the skill reward will be the sum of $T$-step rewards, as shown in Line 32 of Algorithm~\ref{alg:asaprl}. We empirically found our method performed well even without $R_{destination}$ in the highway and roundabout environments.


\begin{figure*}[t]
    \begin{center}
    \includegraphics[width=1\textwidth]{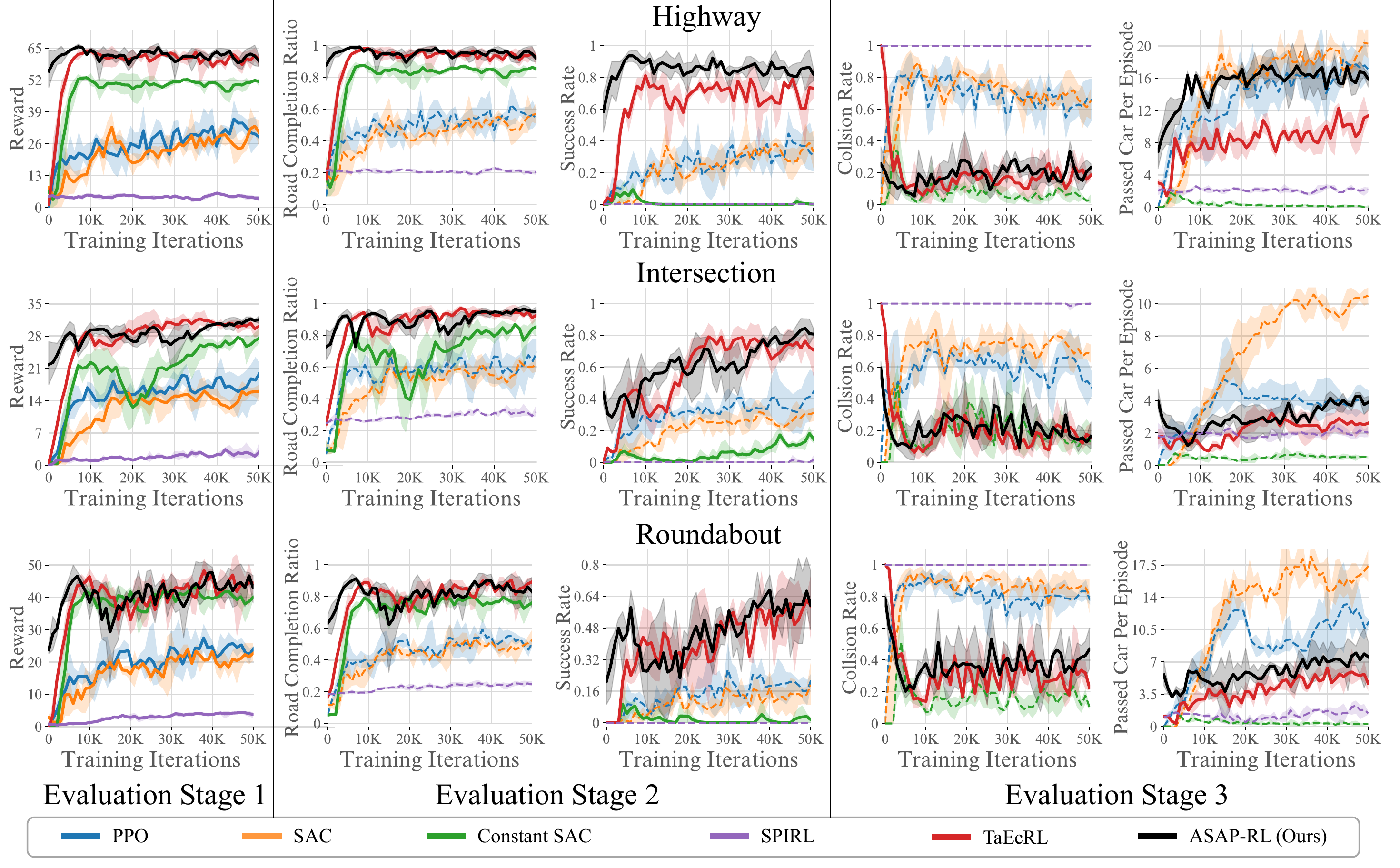}
    \end{center}
    \caption{Comparison of our method with baselines on the highway, intersection, and roundabout scenarios. PPO and SAC are classical RL algorithms over control space. Constant SAC repeats the same action for the skill horizon $T$. SPiRL and TaEcRL learn in low-dimension latent skill space and SPiRL also leverages expert priors. As detailed in Sec.~\ref{exp: evaluation stage}, the performance evaluation follows three stages to gradually distinguish the differences between methods with increasingly more specific metrics: 1) reward; 2) success rate, and road completion ratio; 3) collision rate, and passed car per episode. We only need to inspect metrics in later stages when the methods perform similarly in previous stages. The methods that are outperformed by other methods in previous evaluation stages are marked as dashed lines in later stages.
    Our ASAP-RL outperforms all other methods, and the margin between ASAP-RL and other methods increases as we move from stage 1 to stage 3. }
    \label{fig: main_result}
    \vspace{-1em}
\end{figure*}

    
    
\subsubsection{Expert Demonstration Collection}    
We have two options to collect expert demonstrations, one hand-designed rule-based expert planner, and one trained RL expert agent. The rule-based planner achieved higher performance than the RL expert agent after time-consuming manual tuning. However, the rule-based planner option required more designs in data collection (e.g. perturbance to the expert actions) to collect more diverse and react-to-danger actions, without which the agent trained from the demonstration will fail due to error accumulation~\cite{ross2011reduction}. In comparison, we found the trained RL expert agent can collect high-quality and more diverse demonstrations, and agents trained from such demonstrations achieved stronger performance. Thus we opt for the RL expert agent for data collection. In the future, the expert demonstrations can also be retrieved by human driving data, if available.

\subsubsection{Baselines}
We compare the performance of our ASAP-RL with the following baselines:
\begin{itemize}[leftmargin=*]
 \item \textbf{PPO~\cite{DBLP:journals/corr/SchulmanWDRK17}}: Train an agent from scratch by Proximal Policy Optimization, a typical single-step on-policy algorithm.
 \item \textbf{SAC~\cite{haarnoja2018soft}}: Train an agent from scratch with Soft Actor-Critic, a typical single-step off-policy algorithm.
 \item \textbf{Constant SAC}: Train an agent from scratch with SAC, whose action is repeated $T$ times (the same as the skill horizon of ASAP-RL). This method verifies the performance of simply temporally extending actions from one-step to multi-step without skill design.
 \item \textbf{SPiRL~\cite{pertsch2020accelerating}}: Train an agent with the Skill-Prior RL algorithm, which leverages both skills and priors in the expert demonstration. SPiRL first embeds skills from the demonstration into a low-dimension latent skill space, where the RL agent can learn and explore efficiently.
 SPiRL then trains a skill prior network using the expert demonstrations, which is utilized as a KL divergence term to guide the RL agent and prevent deviation from the expert policy. 
  Since the open-source code of SPiRL does not include the MetaDrive environment, we reproduced SPiRL in our codebase.
 \item \textbf{TaEcRL~\cite{zhou2022accelerating}}: Train an agent with task-agnostic and ego-centric motion skills.
 The difference between TaEcRL and SPiRL lies in a) TaEcRL learns latent skill space through the task-agnostic and ego-centric motion skill library proposed by TaEcRL, while SPiRL directly learns latent space from collected expert data with limited skill diversity; b) SPiRL leverages expert priors but TaEcRL does not. 
\end{itemize}
\subsubsection{Evaluation metric} To compare the performance of ASAP-RL with other methods, we adopt the following metrics, which reflect the performance of autonomous vehicles from different aspects:
\begin{itemize}[leftmargin=*]
 \item Episode Reward: the sum of all the rewards in an episode.
 \item Success Rate: the percentage of episodes where the agent reaches the destination on time without collisions.
 \item Road Completion Ratio: the ratio of road length completed by the agent to the total road length per episode.
 \item Collision Rate: the percentage of episodes in which a collision occurs.
 \item Passed Cars Per Episode: the number of cars overtaken by the agent in each episode.
\end{itemize}

\begin{figure*}[!t]
    \begin{center}
    \includegraphics[width=1\textwidth]{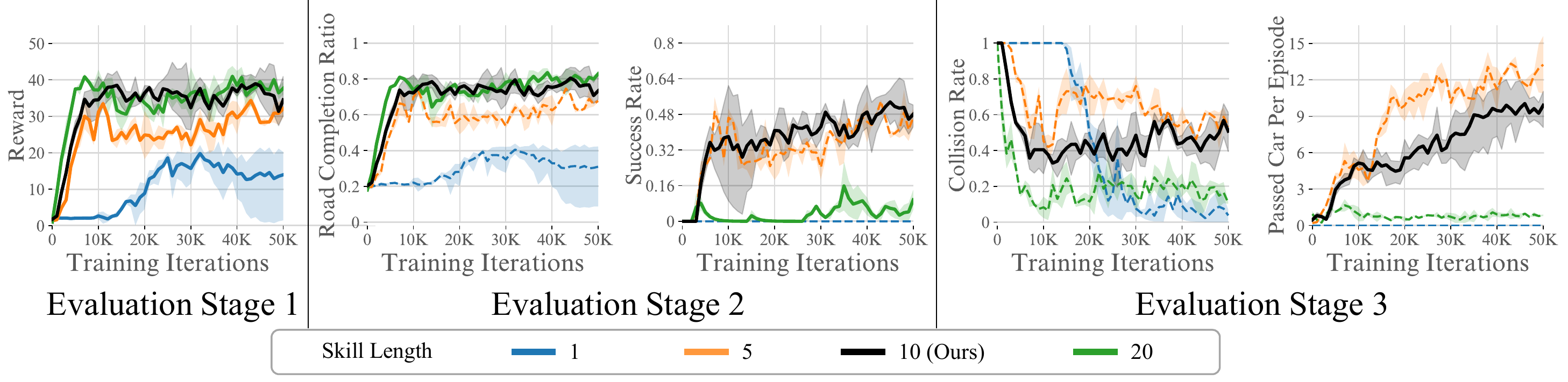}
    \end{center}
    \caption{Ablation analysis of skill length in the roundabout scenario. When $T$ increases from 1 to 10, performance improvement is observed, benefiting from temporal abstraction. But when $T$ reaches 20, it is too long for the agent to react to accidents during the skill execution due to delayed replanning. We observe that a skill length of 10 reached a good trade-off.}
    \label{fig: length}
    \vspace{-1em}
\end{figure*}

\begin{figure*}[!t]
    \begin{center}
    \includegraphics[width=1\textwidth]{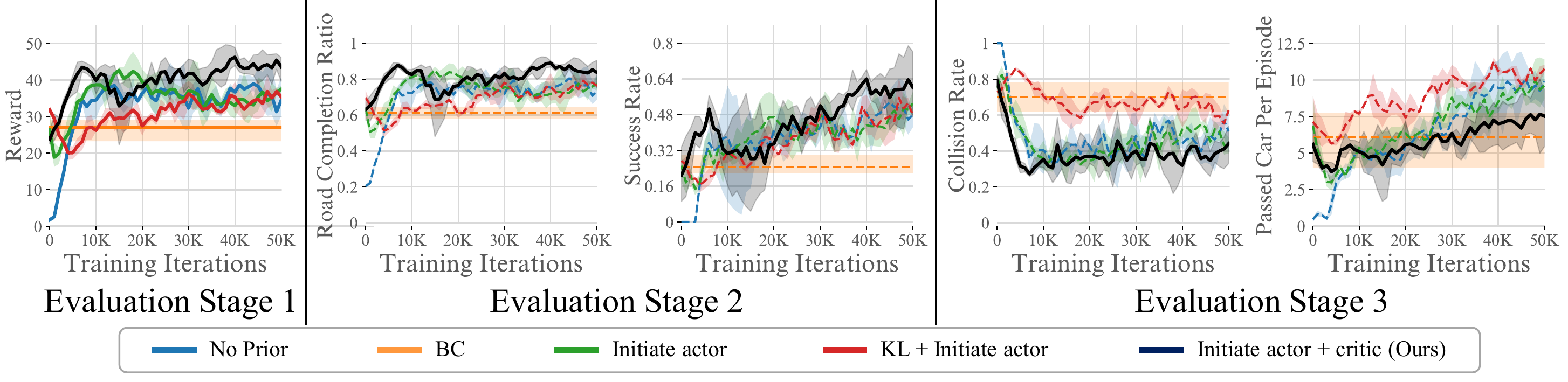}
    \end{center}
    \caption{Ablation analysis of different ways to incorporate expert priors in the roundabout scenario. Our method has a good starting performance and reached the highest final performance without a performance drop in early training iterations ('Initiate actor') or performance suppression due to the expert suboptimality ('KL + Initiate actor').}
    \label{fig: prior}
    \vspace{-1em}
\end{figure*}

\subsubsection{A three-stage strategy for performance evaluation}
\label{exp: evaluation stage}
During the evaluation, the metrics above should be combined and sequentially inspected, since they evaluate the driving performance from different aspects and granularity. For example, a high passed-car-per-episode represents a good agent only when the agent also remains a decent success-rate and road-completion-ratio, and a reasonably low collision-rate. 
Thus we propose a three-stage strategy to evaluate the methods' performance, where we only need to inspect metrics in later stages when the methods perform similarly in metrics of previous stages:
\begin{itemize}
    \item \textbf{Stage one} Since the reward is the direct optimization objective in RL, it can straight-forwardly reflect the performance of an algorithm. Therefore, in the first stage, we treat reward as the most important evaluation metric. 
    \item \textbf{Stage two} Sometimes it can be difficult to distinguish methods on the reward metric alone. In this case, we move to the second stage focusing on other metrics with more driving contexts, namely the success rate, and the road completion ratio. 
    \item \textbf{Stage three} If the performance on previous metrics is still close, we move to the third stage and focus on the collision rate and passed car per episode. The two metrics are somewhat trade-offs between each other. The passed car per episode has the smallest reward weight and can better distinguish the methods' performance. 
\end{itemize}


\subsection{Performance Comparison}
We follow the proposed three-stage strategy to evaluate the performance of baselines and our method. As illustrated in Fig.~\ref{fig: main_result}, the reward metric in the first evaluation stage indicates that the performance of PPO, SAC, and SPiRL is significantly lower than that of the other three methods. \citet{zhou2022accelerating} reported that PPO and SAC had poor performance in MetaDrive tasks even under dense reward settings, so it is not surprising that the performance is poor under the more difficult sparse reward conditions used in this work. Though SPiRL was shown to be effective in manipulation tasks~\cite{dalal2021accelerating,pertsch2020accelerating}, it performs poorly in our driving setting, likely due to the fact that the driving task requires very diverse skills while SPiRL can only learn skills from limited expert demonstrations: since the expert demonstrations of driving tasks are usually unbalanced to mostly consist of data where the vehicle is driving forward, it is hard to ensure that all essential skills are learned or covered in SPiRL agent. We empirically found that the agents learned with SPiRL mostly drive straight forward and can barely exhibit maneuvers, leading to a high collision rate.
 
The constant-SAC, which simply repeats the same action in the skill window $T$, obtains a slightly lower reward in the three scenarios compared to TaEcRL and ASAP-RL. However, when we move to the second/third evaluation stage and focus on the detailed driving-context metrics, the Constant-SAC performed much worse than TaEcRL and ASAP-RL:
the constant-SAC agent drives very conservatively and slowly to avoid collisions. Though a low collision rate and a decent road completion ratio are achieved, the agent usually cannot arrive at the destination within the given time, which leads to a close-to-zero success rate and passed cars per episode. Thus it is not a good driving strategy and shows the importance of proper skill design.

Finally, we compare the results between TaEcRL and ASAP-RL. ASAP-RL achieves better performance in the first 10k iterations in terms of all metrics due to the usage of the expert prior. In addition, ASAP-RL has more passed cars per episode than TaEcRL by a large margin, with better or similar success rates, road completion ratio, and collision rates. This is hypothetically because 1) our ASAP-RL can generate very diverse skills by combining different skill parameters, while TaEcRL relies on offline datasets and can only use limited skills. For example, we found agents learned by TaEcRL usually drive in an erratic way, leading to a low ability to overtake. 2) ASAP-RL can leverage expert priors to accelerate convergence and bias the agent toward better optima and higher performance, due to our proposed skill parameter recovery and double-initialization method. In contrast, TaEcRL is limited by its own algorithm design and cannot utilize expert data. These results suggest ASAP-RL could be a useful tool to enable autonomous driving in interactive, diverse dense-traffic scenarios.



\begin{figure*}[t]
    \begin{center}
    \includegraphics[width=1\textwidth]{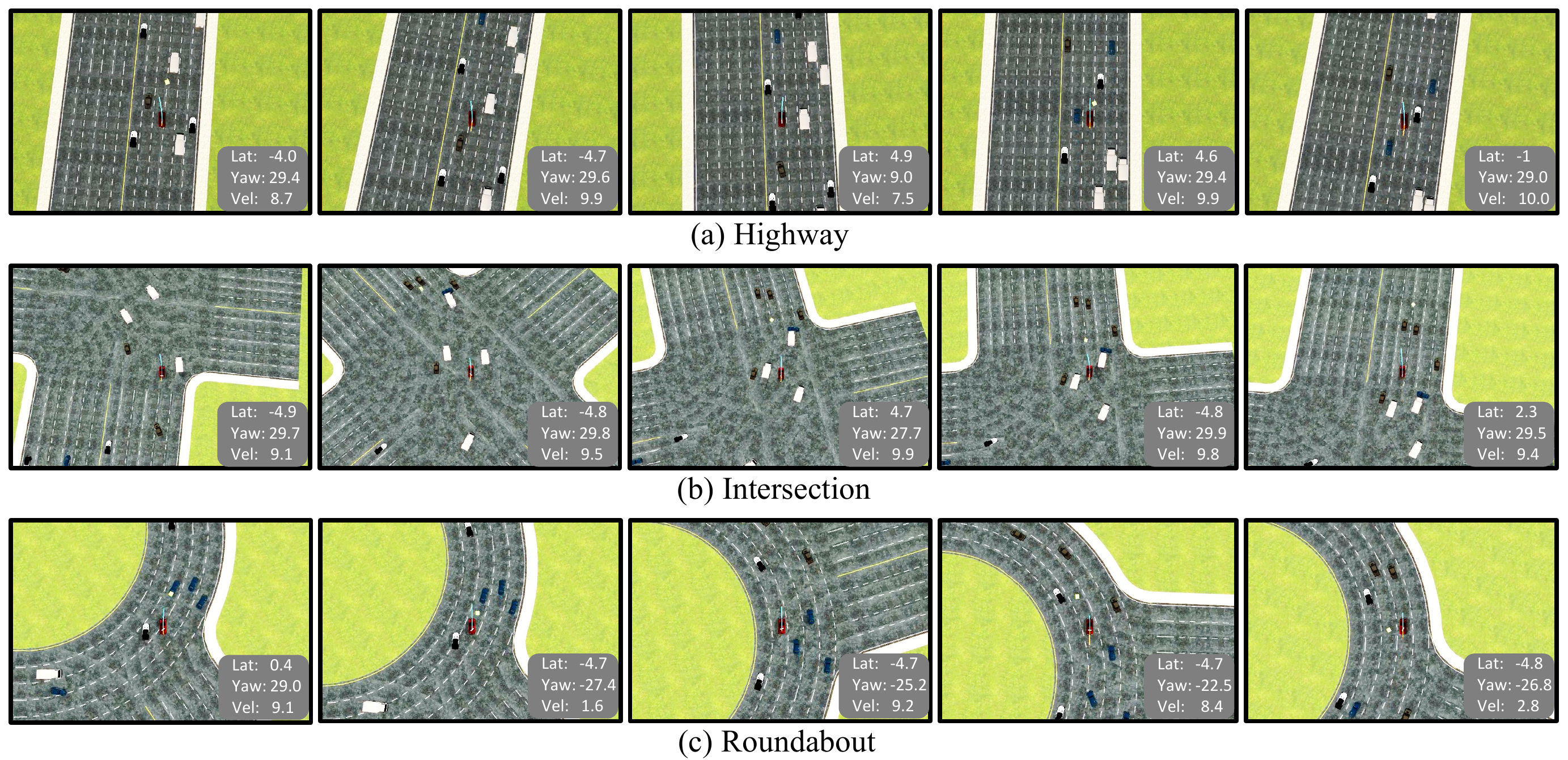}
    \end{center}
    \caption{Visualizations of the trained agents in three dense-traffic scenarios. The dark red vehicle denotes the ego agent and the blue line denotes the motion skill trajectory output by the agent. Generated parameters corresponding to the skill are shown in the bottom right. (a) In the highway scenario, the agent performs consecutive lane changes to overtake vehicles ahead. (b) In the intersection scenario, the agent turns left and overtakes vehicles ahead with a high velocity. (c) In the roundabout scenario, the agent slows down to cautiously overtake the vehicle on the left (the second image), and drives slowly when vehicles ahead block the way (the last image).}
    \label{fig: case study}
    \vspace{-1em}
\end{figure*}

\subsection{Ablation study}
\subsubsection{Influence of the length of skill}
We ablate the skill horizon parameter, $T$, on the roundabout scenario to examine its influence. The results are shown in Fig.~\ref{fig: length}. When $T=1$, ASAP-RL degenerates into a single-step method with poor performance. When $T=5$ or $T=10$, we observed a performance improvement benefiting from making long-term decisions. However, if $T$ is too large ($T=20$), during the skill execution, an overly-long skill length can make the agent less reactive to accidents and emergencies due to delayed replanning. Moreover, the richness of the skill set will be insufficient, and thus some driving strategies cannot be covered. Overall, a skill length of $T=10$ reached a good trade-off.

\subsubsection{Influence of expert prior}
\label{abl: prior}
We conduct ablation studies on the roundabout scenario to investigate the effect of the expert prior and compare the proposed double initialization method with other methods to incorporate the expert prior. 
As shown in Fig.~\ref{fig: prior}, we can distinguish their differences simply using the reward metric in the first evaluation stage.

\noindent{\textbf{No Prior.}} No Prior has to learn from scratch with low performance at the beginning with close-to-zero rewards, high collision rates, and data collection costs. Though its performance then goes up quickly, it fails to improve further. For example, in the reward metric, No Prior is constantly lower than our ASAP-RL at any training iteration, and the peak reward of our ASAP-RL is 20\% higher than that of No Prior. These results demonstrate the effect of priors to bias agents toward better optima and higher performance.

\noindent{\textbf{Behavior Cloning (BC).}} The BC method trains a policy using expert demonstration without further reinforcement learning. Technically, the training is the same as the actor pretraining in our method as in Eq~\ref{eq:ELBO}. Compared to 'No Prior', the BC method achieves better performance at the beginning but does not improve further, so it has a lower final performance.

\noindent{\textbf{Initiate Actor.}} The 'Initiate Actor' method uses the pre-trained policy weight to initialize the actor in SAC, and then continue with reinforcement learning. This method has a good initial performance and continues to improve. However, there is a performance drop in the first 5K iterations, likely due to the mismatch between the actor and the critic: though the actor is pretrained to have expert prior knowledge, the critic has no knowledge of the prior. Because the actor needs to interact with the critic during RL with the objective to maximize the Q-value output by the critic, as in Line 37 of Algorithm~\ref{alg:asaprl}, the actor could quickly lose the prior knowledge learned from the expert after several updates.

\noindent{\textbf{KL + Initiate Actor.}} In addition to actor initialization, the `KL + Initiate Actor' method also measures the KL divergence between the pre-trained actor and the RL actor during training. The KL term was added into the objective function to replace the original entropy term in SAC as in ~\cite{pertsch2020accelerating}. This is an ablation setting that incorporates expert prior during the whole RL training to guide the RL and prevent deviating from the expert policy. However, similar to the results reported in ~\cite{rengarajan2022reinforcement}, the performance of the method is suppressed due to expert suboptimality: though this method has a good starting point, the reward is growing much more slowly than other methods in later training.

\noindent{\textbf{Initiate Actor and Critic.}} This is our proposed method to incorporate expert priors.  Our `double initialization' method has a good starting reward and the highest final performance, without the issue of performance drop caused to a mismatch between actor and critic, or performance suppression due to the expert suboptimality. 





\subsection{Visualizations}
Visualizations of our ASAP-RL agent running on the three scenarios can be found in Fig.~\ref{fig: case study}. The agent drives in dense traffic efficiently and safely with diverse maneuvers such as lane changing, cutting in, braking, etc.

\section{Conclusion}
We present an efficient reinforcement learning (ASAP-RL) that simultaneously leverages parameterized motion skills and expert priors for autonomous vehicles to navigate in complex dense traffic. We first introduce parameterized motion skills and enable RL agents to learn over the skill parameter space instead of the control space. To further leverage expert priors on top of skills, we propose an inverse skill parameter recovery technique to convert expert demonstrations from control space to skill space. A simple but effective double initialization technique is also introduced to better leverage expert priors. Validations on three challenging dense-traffic driving scenarios demonstrate that our ASAP-RL significantly outperforms previous methods in terms of learning efficiency and performance.




\bibliographystyle{plainnat}
\bibliography{references}

\clearpage

\end{document}